\documentclass[conference]{IEEEtran}
\IEEEoverridecommandlockouts
\usepackage{cite}
\usepackage{amsmath,amssymb,amsfonts}
\usepackage{algorithmic}
\usepackage{graphicx}
\usepackage{textcomp}
\usepackage{xcolor}
\usepackage{multirow}
\usepackage{diagbox}
\usepackage[switch]{lineno} 
\usepackage{threeparttable} 
\def\BibTeX{{\rm B\kern-.05em{\sc i\kern-.025em b}\kern-.08em
    T\kern-.1667em\lower.7ex\hbox{E}\kern-.125emX}}
\begin{document}

\title{DADNN: Multi-Scene CTR Prediction via Domain-Aware Deep Neural Network\\
\thanks{[a]These authors have contributed equally to this work.}
}

\author{ Junyou He$^{1,a}$,Guibao Mei$^{1,a}$,Feng Xing,Xiaorui Yang,Yongjun Bao,Weipeng Yan\\ Business Growth BU, JD\\ \{hejunyou1,meiguibao,xingfeng7,lucky.yang,baoyongjun,Paul.yan\}@jd.com\\}

\maketitle

\begin{abstract}
Click through rate(CTR) prediction is a core task in advertising systems. The booming e-commerce business in our company, results in a growing number of scenes. Most of them are so-called long-tail scenes, which means that the traffic of a single scene is limited, but the overall traffic is considerable. Typical studies mainly focus on serving a single scene with a well designed model. However, this method brings excessive resource consumption both on offline training and online serving. Besides, simply training a single model with data from multiple scenes ignores the characteristics of their own. \\
\indent To address these challenges, we propose a novel but practical model named Domain-Aware Deep Neural Network(DADNN) by serving multiple scenes with only one model. Specifically, shared bottom block among all scenes is applied to learn a common representation, while domain-specific heads maintain the characteristics of every scene. Besides, knowledge transfer is introduced to enhance the opportunity of knowledge sharing among different scenes. In this paper, we study two instances of DADNN where its shared bottom block is multilayer perceptron(MLP) and Multi-gate Mixture-of-Experts(MMoE) respectively, for which we denote as DADNN-MLP and DADNN-MMoE. MMoE is adopted to model the scene differences while still capturing the commonness, without requiring significantly more model parameters compared to MLP. Comprehensive offline experiments on a real production dataset from our company show that DADNN outperforms several state-of-the-art methods for multi-scene CTR prediction. Extensive online A/B tests reveal that DADNN-MLP contributes up to 6.7\% CTR and 3.0\% CPM(Cost Per Mille) promotion compared with a well-engineered DCN model. Furthermore, DADNN-MMoE outperforms DADNN-MLP with a relative improvement of 2.2\% and 2.7\% on CTR and CPM respectively. More importantly, DADNN utilizes a single model for multiple scenes which saves a lot of offline training and online serving resources.
\end{abstract}


\section{Introduction}
\indent As one of the largest B2C e-commerce platforms in China, our company serves hundreds of millions of active customers. Accurate CTR prediction is the cornerstone of the entire advertising system, directly related to the use experience of hundreds of millions of active users of our company and the vital commercial interests of advertisers. Recently, inspired by the success of deep learning in many research fields such as computer vision\cite{2017Densely} and natural language processing\cite{2014Neural}, the deep learning for CTR prediction\cite{7837964} has attracted growing attention from both academia and industry. Some neural network based models have been proposed and achieved great success, such as Wide\&Deep model(WDL)\cite{2016Wide}, DCN\cite{2017Deep}, DeepFM\cite{2017DeepFM}, xDeepFM\cite{2018xDeepFM} etc.
However, the previous methods are proposed for a single scene.
In the large scale online platform of our company, there are lots of scenes whose traffic are various. Some long-tail scenes of the online advertising are illustrated in Fig. \ref{fig:scene_examples}.
There are several challenges using existing approaches to model the CTR prediction in these scenes:
\begin{figure}
	\centering
	\includegraphics[width=0.4\textwidth]{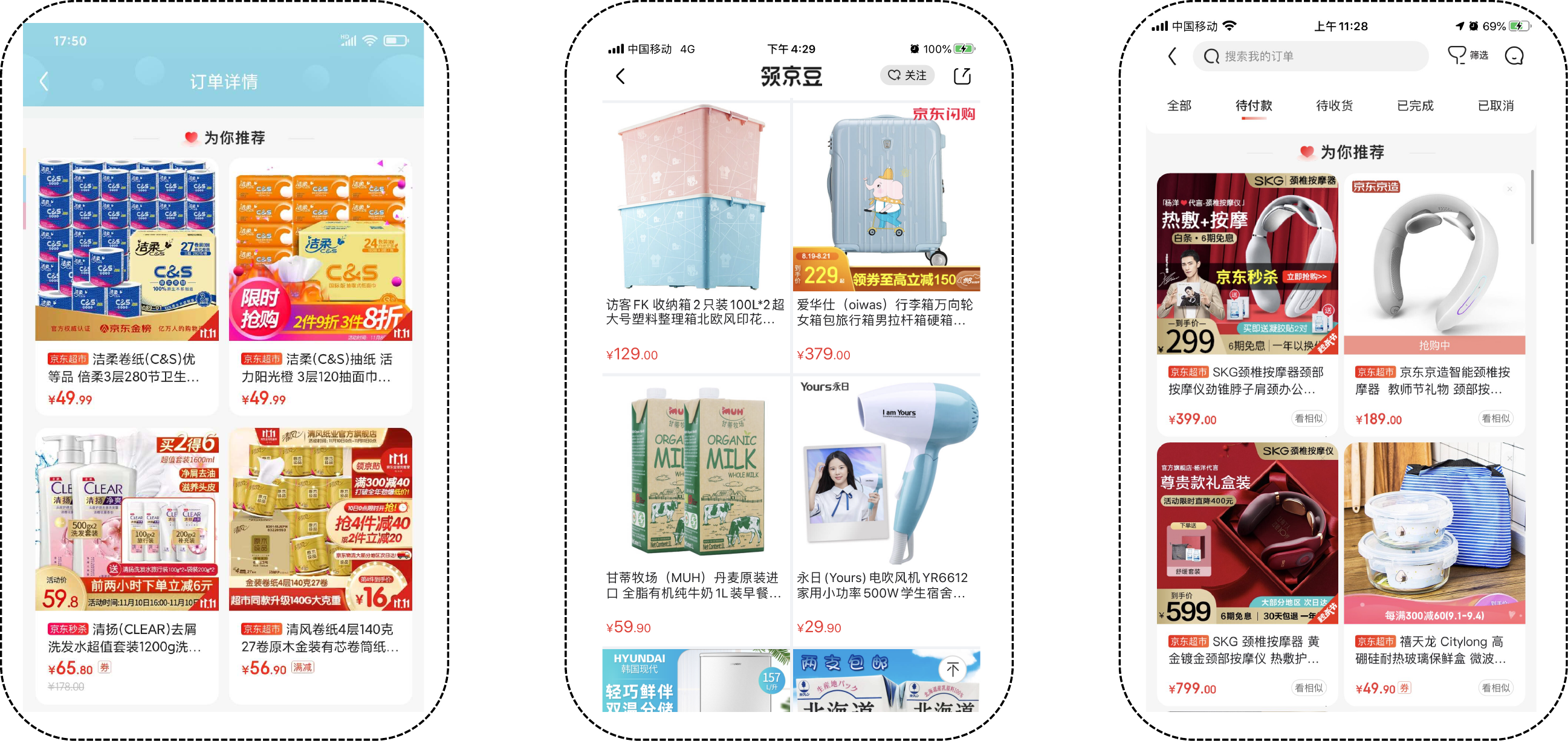}\\
	\caption{Scene examples.}\label{fig:scene_examples}
\end{figure}
\begin{enumerate}
	\item The scene with small volume i.e. insufficient samples is not enough to support an individual model optimization while having a separate model for each scene is also often prohibitive in real production settings due to limited offline training and online serving resources. Moreover, parameter tuning for each scene is labor-intensive.
	\item The data distributions of individual scenes are usually different. Training a model such as DCN or DeepFM by samples from multiple scenes may suffer from domain shift problem, thus generally performing poorly.
\end{enumerate}
\indent In this case, there is an urgent need for an efficient way to model those scenes simultaneously. To address above challenges, we propose a novel model named Domain-Aware Deep Neural Network(DADNN) for multi-scene CTR prediction. More specifically, as far as we know the data distribution of each scene is different and the data from multiple scenes are likely complementary. Hence, we introduce a shared-bottom block to learn a generic representation by leveraging all data over multiple scenes. Following the shared bottom block, the domain-specific heads learn the discriminative characteristics of each individual scene, thus reducing the domain shift.
In addition, inspired by Multi-gate Mixture-of-Experts(MMoE)\cite{2018Modeling}, which explicitly learns to model task relationships from data, we introduce it for another purpose. MMoE module in our model helps to automatically determine the importance of the shared representations for the respective scene, allowing both scene-shared and scene-specific representations to be learned.
Finally, we believe that the knowledge of different scenes can be learned from each other under the same task. Thus, we go one step further to promote interactions across scenes through knowledge transfer, whilst it also help alleviate insufficient training of the domain specific heads in scenes with limited samples.
It is worth mentioning that the model is extensible to continuously support new scenes. \\
\indent The contributions of this work are summarized as follows:
\begin{enumerate}
	\item To the best of our knowledge, this is a first study using a transfer learning (TL) model for multi-scene CTR prediction. The proposed method can support multiple scenes with a single model, thus greatly saving human labor and computation resources(offline training and online service). It is also scalable to continuously support new scenes.
	\item In order to minimize the domain shift in different scenes, we propose a routing and domain layer where each scene has an individual domain-specific head similar to the multi-task learning (MTL) framework. Different from the traditional multi-task learning application, we solve the same task in different scenes which share the underlying representation. 
	\item Considering that the training of domain-specific heads is insufficient in scenes with limited samples and the knowledge of different scenes may be complementary under the same task, we adapt knowledge transfer (KT) among multiple scenes to enhance the opportunity of knowledge sharing by an internal KT process limited in a single model. We shift KT primary focus away from utilizing an extra teacher net to get compact models.
	\item There is a certain difference among scenes. In the shared-bottom block, if the differences of scenes can be captured explicitly while considering the commonality, the CTR prediction performance can be further boosted. To this end, we introduce MMoE module. Each expert captures the commonness of samples from different standpoints, and the weights of gate allow for discriminative representations to be tailored for each individual scene. It is worth noting that the shared bottom network could be any popular models, not limited to MMoE.
	\item DADNN has been successfully deployed in the online advertising system of JD, one of the largest B2C e-commerce platform in China. We have also conducted online A/B test to evaluate its performance in real-world CTR prediction tasks and obtained significant improvement in terms of CTR and CPM respectively.
\end{enumerate}
\indent This paper may open up a new idea for multi-scene CTR prediction of advertising. We believe that more advanced modeling techniques can be integrated into our model, which may further improve performance.\\
\indent The rest of this paper is organized as follows: Section 2 reviews related works. Section 3 introduces the proposed approach in details. Section 4 evaluates the proposed models on the dataset of our company. Finally, we conclude the paper in Section 5.

\section{Related Work}
Our work is closely related to three research areas: CTR prediction, multi-task learning and transfer learning.

\subsection{CTR Prediction}
The goal of CTR prediction is to predict the probability that a user will click on an ad. CTR prediction has attracted the attention of many researchers from both academia and industry. In recent years, Deep Neural Networks (DNNs) have shown powerful ability of automatically learning feature representations and high-order feature interactions\cite{2017Densely,2014Neural}, which are exploited for CTR prediction. Representative models include Wide\&Deep\cite{2016Wide}, DeepFM\cite{2017DeepFM}, and DCN\cite{2017Deep}. Wide\&Deep \cite{2016Wide} jointly trains Logistic Regression (LR) and DNN to improve both the memorization and generalization abilities for recommender systems. DeepFM \cite{2017DeepFM} replaces the wide part of Wide\&Deep with FM and shares the feature embedding between the FM and DNN, which further improves the model ability of learning feature interactions. DCN\cite{2017Deep} develops a multi-layer residual structure to learn higher-order feature interactions efficiently in an explicit fashion. However, these approaches only consider a single scene for CTR prediction.

\subsection{Multi-task Learning}
\indent Multi-task learning is a hot research topic in the deep learning field that aims at improving the generalization performance of a task using other related tasks. Multi-task models can learn commonalities and discrimination across different tasks. It is necessary to model multiple objectives in real-world applications, such as optimizing CTR and CVR simultaneously in recommender or advertisement systems. The most commonly used multi-task DNN structure learns the general representation through a shared-bottom module\cite{1998Multitask}.
Such framework has been widely used in online advertising \cite{2017Deep,2018Entire} to solve multiple objectives. Instead of hard parameter sharing across tasks, some recent works consider constraints on task-specific parameters. Hinton\cite{Jacobs1991Adaptive} proposes One-gate Mixture-of-Experts(OMoE) where the shared bottom layer is divided into several sub-networks (called experts), and a gate is set up to make different data use the sharing layer in a diversified way compared with the general multi-task learning framework. Further, MMoE\cite{2018Modeling} splits the shared low-level layers and uses different gating networks for different tasks to utilize multiple sub-networks. Our method is partially inspired by these prior works, however, the most significant difference is that we aim at a single task,namely,multi-scene CTR prediction. Moreover, the added routing layer distinguishes scenes by a wide input of scene id and one domain layer only use single scene data. We shift MTL primary focus away from learning multiple tasks simultaneously.
\begin{figure*}[t]
	\centering
	\includegraphics[width=0.95\textwidth]{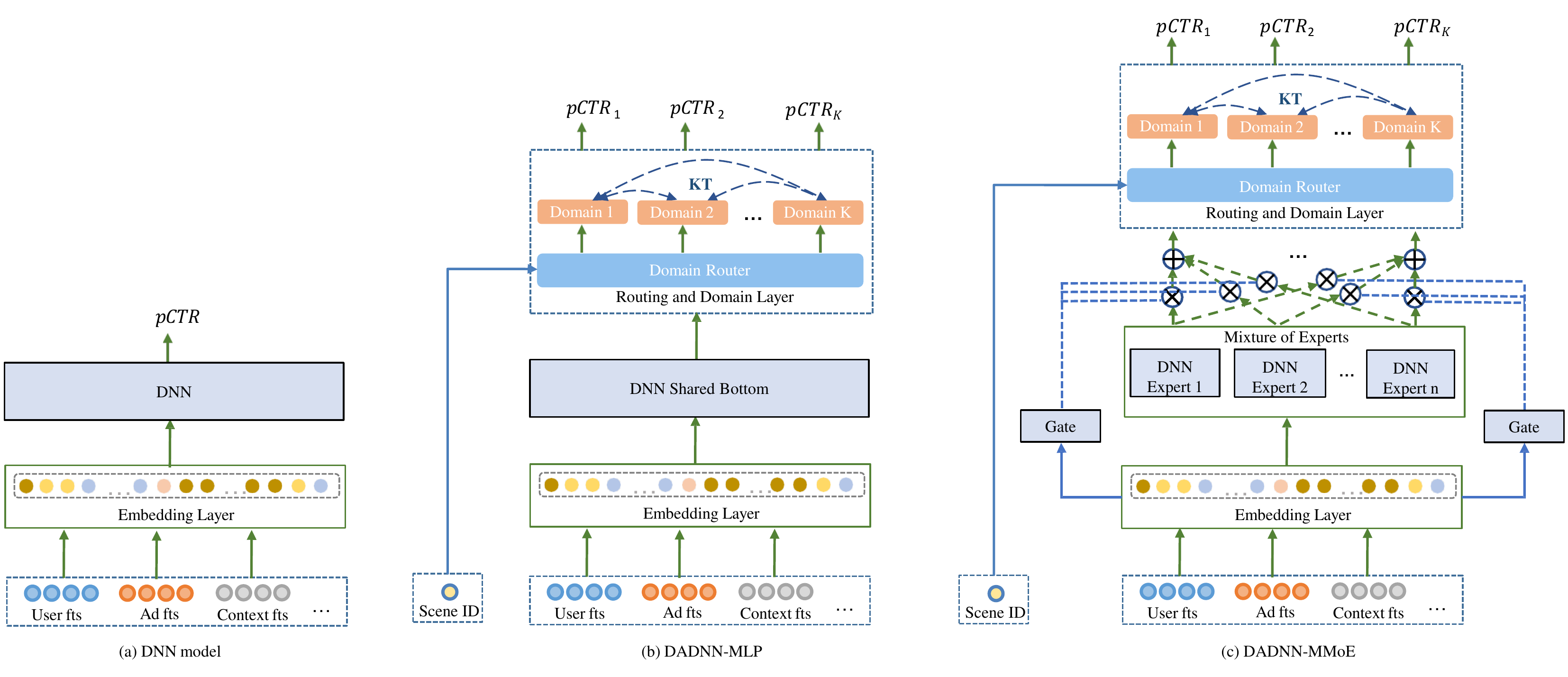}\\
	\caption{Illustration of model architectures (fts-features). (a) DNN model, which considers only a single scene. (b) DADNN-MLP model, which further considers discriminative characteristics to be tailored for each individual scene. The routing layer uses a wide input of the scene id to distinguish the scene. KT represents internal knowledge transfer among multiple scenes. (c) DADNN-MMoE model, which introduces the multi-gate mixture-of-experts to substitute the hard shared-bottom block. The weights of gate allow for discriminative representations for each individual scene.}
	\label{fig:model_architectures}
\end{figure*}
\subsection{Transfer Learning}
\indent Transfer learning tries to solve the basic problem of insufficient training data. Domain adaptation (DA) is a particular case of transfer learning (TL) that utilizes knowledge distilled from a source domain to execute new tasks in a target domain which has a different but related data distribution. Recently, a number of deep domain adaptation approaches have been proposed to solve a domain shift between the source and target domains. Long et al. \cite{2015Learning} proposes the deep adaptation network (DAN) to reduce distribution shift across domains by adding multiple adaptation layers and exploring multiple kernels. Domain adversarial neural network (DANN) \cite{Ganin2014Unsupervised} utilizes a shared feature extraction layers and two classifiers to reduce the domain discrepancy by a minimax game. Conventional DA focuses on high-performance target domain learner. In contrast, we utilize domain towers to reduce domain shift in multiple scenes, which are both source and target domains. Moreover, our proposed knowledge transfer can be viewed as an internal knowledge transfer process limited in one single model among multiple scenes, where the data distributions may be different. Domain adaptation has garnered less attention for the CTR prediction tasks.

\section{Model Architecture}
\indent Classical CTR prediction models focus on a single scene(illustrated in Fig. \ref{fig:model_architectures}(a)), thus the performance may be degraded due to domain shift when serving multiple scenes. To this end, we propose the Domain-Aware Deep Neural Network(DADNN) for multi-scene CTR prediction in Fig. \ref{fig:model_architectures}(b) and Fig. \ref{fig:model_architectures}(c). In this section, we introduce our model in detail. The proposed model is composed with four parts:i) Sparse Input and Embedding Layer, ii)Shared-Bottom Block,   iii)Routing and Domain Layer, iv)Knowledge Transfer.

\subsection{Sparse Input and Embedding Layer}
\indent The sparse input layer and embedding layer are widely used in most of the popular model structures such as DeepFM and DCN. Each ad instance can be described by multiple fields such as user profile("Age","Gender",etc.), ad profile("ad ID","ad Tags",etc.) and context profile("Network Type",etc.). The instantiation of a field is a feature. The sparse input layer transforms raw input features into high-dimensional sparse binary features. The embedding layer is able to transform these binary features into low-demensional dense vectors. Through embedding, we obtain $f$ embedding vectors ${\mathbf{e}_i \in R^D,}i=0,\cdots,f$. Then all the vectors are concatenated to obtain the overall representation vector for the instance:$\mathbf{E} = [\mathbf{e}_1, \mathbf{e}_2,\cdots  , \mathbf{e}_i , \cdots , \mathbf{e}_f ]$, where $\mathbf{e}_i$ is the embedding of $i$-th field and $D$ is the embedding size. In practice, the embedding layer is pre-trained using a large-scale scene data to further improve the performance of DADNN. Reusing the embedding layer helps converge faster and better and get a more expressive representation.

\subsection{Shared-Bottom Block}
\indent As far as we know, feeds have a certain similarity. It can be considered that the information of multiple scenes is complementary. We introduce a shared-bottom block to learn a generic feature representation by leveraging all data over multiple scenes. There are many choices of the shared-bottom block. We implement two natural types: the first one, which we call hard shared-bottom block (shown in Fig. 2(b)), consists of several large low-level layers that are shared by all scenes; the second one, which we call MMoE (shown in Fig. 2(c)), splits the shared low-level layers into sub-networks called experts.
In the former case, the shared network, represented as function $f(\mathbf{x})$, learns a compact global feature pool across all scenes, after which a set of domain-specific heads predict the CTR for each scene. The output $y_k$ for $k$-th scene follows the corresponding domain-specific head $h^k$:
\begin{equation}
	y_k = h^k(f(\mathbf{x})).
\end{equation}
\indent MMoE further utilizes a group of experts instead of hard shared-bottom block. Moreover, a gating network for each scene ensembles the results from all experts,  which is similar to attention essentially. Through introducing the MMoE module, our model could explicitly capture the scene differences without requiring significantly more model parameters than the hard shared-bottom model. We use a separate gating network $g_k$ and $n$ experts for each scene $k$. The output $y_k$ of scene $k$ is obtained as follows:
\begin{equation}
	y_k = h^k(f^k(\mathbf{x})),
\end{equation}
where   
\begin{equation}
	f^k(\mathbf{x}) = \displaystyle \sum_{j=1}^n g^k(\mathbf{x})_j f_j(\mathbf{x}).
\end{equation}
The gating networks are simply linear transformations of the input with a softmax layer:
\begin{equation}
	g^k(\mathbf{x}) = softmax(\mathbf{W}_{gk}\mathbf{x}),
\end{equation}
where $\mathbf{W}_{gk}$ is trainable.
Following the shared-bottom block, the routing layer splits all scene data into different domain layers.

\subsection{Routing and Domain Layer}
The data of individual scenes rarely, if ever, follow identical distribution.
In order to reduce domain shift across scenes, the routing layer splits the samples by scenes into respective domain layer, thus allowing for discriminative representations to be tailored for each individual scene. The routing layer distinguishes scenes by a wide input of scene id. When serving online, only one domain layer is activated for each scene. The routing and domain layer is shown in Fig. 2(b) and Fig. 2(c).
More specifically, each scene has a domain layer which only uses its own data to adjust model parameters. To this end, domain layer can relieve the degradation in performance of introducing multiple data distributions. Given a dataset $D=\{(\mathbf{x}_i,y_i)|i=1,2,...,N\}$, the objective function of our model is defined as follows:
\begin{equation}
	\mathop{\arg\min}_{\mathbf{W}_d}L_d(\mathbf{W}_d;D),
\end{equation}
where $L_d$ is the total loss over training set. It can be formulated as:
\begin{equation}
	L_d(\mathbf{W}_d;D) = \sum_{k=1}^K \alpha_k L_{d_k},
\end{equation}
where $L_{d_k}$ is the loss for the $k$-th scene with $\alpha_k$ as the corresponding weight and $K$ is the number of scenes. Through our exploration, we find that it works well when setting $\alpha_k$ to be the percentage of $k$-th scene's samples dynamically.
Specifically, $L_{d_k}$ is typically defined as the cross-entropy loss function:
\begin{equation}
	L_{d_k} = - \frac{1}{N_k}\sum_{i=1}^{N_k}(y_ilogp(\mathbf{x}_i)+(1-y_i)log(1-p(\mathbf{x}_i))),
\end{equation}
where $N_k$ is the size of $k$-th scene samples, $y_i$ is the ground truth of $i$-th instance, and $p(\mathbf{x}_i)$ is the output of the $k$-th domain layer which represents the probability of sample $\mathbf{x}_i$ being clicked.

\subsection{Knowledge Transfer}
\indent Although the routing and domain layer could alleviate the domain shift, the domain layer for scenes with limited traffic may be insufficiently trained. Besides, in our experiment, these scenes are certain similar feeds. To this end, we propose a knowledge transfer module between every two scenes to enable comprehensive knowledge interactions and information sharing, which is shown in Fig. 2(b) and Fig. 2(c). Once these knowledge from teacher domain classifier is generated, it can guide other domain layers by another cross-entropy loss. Below, we describe the mathematical formulation of our proposed knowledge transfer method.
Given a dataset $D=\{(\mathbf{x}_i,y_i)|i=1,2,...,N\}$, the objective function of our model is defined as follows:
\begin{equation}
	\mathop{\arg\min}_{\mathbf{W}_d,\mathbf{W}_{kt}}L_d(\mathbf{W}_d;D)+L_{kt}(\mathbf{W}_{kt};D).
\end{equation}
Specifically, $L_{kt}$ is knowledge matching loss which represents the pairwise probabilistic prediction mimicking loss extended from \cite{li2020dynamic, hinton2015distilling}, which is defined as:
\begin{equation}
	L_{kt} = \sum_{p=1}^K \sum_{\mbox{\tiny$\begin{array}{c}
				q=1 \\
				p \neq q
			\end{array}$}} ^K u_{pq}L_{pq},
\end{equation}
\begin{equation}
	L_{pq} = -\frac{1}{N_p}\sum_{i=1}^{N_p} (p(\mathbf{x}_i)logq(\mathbf{x}_i)+(1-p(\mathbf{x}_i))log(1-q(\mathbf{x}_i))),
\end{equation}
where $p(\mathbf{x})$ and $q(\mathbf{x})$ represent teacher network and student network respectively.  $u_{pq}$ is the corresponding weight of classifier p to q and $N_p$ is the size of teacher samples. In our experiment, we set the $u_{pq}$ to be 0.03. In particular, we update the student network only with the scene data in the teacher network. We develop the gradient block scheme to prevent the teacher net from deterioration, which is exploited in \cite{Zhou2018RocketLA}.

\section{Experiment}
\indent In this section, we present our experiments in detail. We conduct experiments on an industrial dataset collected from the online advertising system in our company to compare DADNN with the state of the art. Furthermore, we design experiments to verify the effect of routing and domain layer, MMoE module and knowledge transfer, respectively. At last, we share the results and techniques for online serving.

\subsection{Metrics}
\indent Area under receiver operator curve(AUC)\cite{2005An} is used commonly in the CTR prediction field. It reflects the probability that a CTR predictor ranks a randomly chosen positive item higher than a randomly chosen negative item. As the data distributions of individual scenes are usually various, we use a metric deriving from GAUC \cite{2017Deep}, which represents a weighted average of AUC calculated in the samples group by each scene. An impression based GAUC is calculated as follows:
\begin{equation}
	GAUC = \frac{\sum_{i=1}^K {impression}_i \times AUC_i}  {\sum_{i=1}^K {impression}_i},
\end{equation}
where the weight is the impression number of each scene. This metric measures the goodness of intra-scene order and is shown to be more relevant to online performance in advertising system. In particular, we use the calibration metric to measure the model stability since accurate prediction of CTR is essential to the success of online advertising. It is the ratio of the average estimated CTR and empirical CTR:
\begin{equation}
	calibration= pCTR/CTR.
\end{equation}
The less the calibration differs from 1, the better the model is.

\subsection{Datasets and Experimental Setup}
\begin{table}
	\centering
	\caption{Statistics of experimental dataset}
	\begin{threeparttable}
		\begin{tabular}{|c|c|c|c|c|c|}
			\hline
			Scene &User(M) & Ad(M) & Samples(M) & CTR\tnote{*} \\
			\hline
			1 &1.3	&1.6		&29.9&  3.63x  \\
			\hline
			2 & 23.5	&2.7		&122.0 & 1.53x   \\
			\hline
			3 & 3.5	&1.4		&40.5 & 1.93x  \\
			\hline
			4 &4.5	&0.4		&16.2 & 9.66x  \\
			\hline
			5 & 18.5	&2.4		&77.5 & x  \\
			\hline
			6 &24.1	&6.2		&184.1 & 2.34x  \\
			\hline
			all &45.7	&7.6		&470.8 & 2.21x  \\
			\hline
		\end{tabular}
		\begin{tablenotes}    
			\footnotesize 
			\item[*] Because of commercial concerns, we only report the relative CTR values
		\end{tablenotes}  
	\end{threeparttable}
	\label{tab:industrial_dataset}
\end{table}

\indent To our knowledge, there is no suitable public dataset for our experiment. Thus, the experimental dataset is collected from traffic logs in our company. We use sampled data in 15 days for training and samples of the following day for testing. The size of training and testing set is about 470.8 millions and 33 millions respectively. The statistics of the above dataset is shown in Table \ref{tab:industrial_dataset}. Because of commercial concerns, we only report the relative CTR values. From the Table \ref{tab:industrial_dataset}, we can see that the volume of all scenes is large and the information of users together with ads is richer than that in a single scene. It is also observed that CTRs of multiple scenes are quite different, which brings even more challenges.
For all deep models, the dimensionality of embedding vectors for each feature is 5. we set the mini-batch size to be 3000, which is proportionally populated by the samples from the six scenes. The optimizer is Adagrad and the learning rate is set to 0.015. For the models except the DADNNs, layers of MLP is set by $200\times 200\times200$. The layers of shared bottom block is set by $200\times 200\times200$ for DADNN-MLP and $100\times 100\times100$ with two experts for DADNN-MMoE. Each domain layer is set by $200\times1$ for DADNN-MLP and $100\times1$ for DADNN-MMoE to require fewer parameters for per scene.

\subsection{Experiments on Industrial Dataset}\label{AA}
\begin{table*}
	\centering
	\caption{Test AUC, calibration and GAUC on industrial dataset}
	\begin{tabular}{|c|c|cccccc|c|}
		\hline
		Method &\diagbox{Metric}{Scene}& 1 & 2 & 3 & 4 & 5 & 6 & GAUC \\
		\hline
		\multirow{2}*{DNN(single)} & AUC & 0.69734 & 0.73114 & 0.74576 &0.66172 & 0.78065 & 0.66585 &\multirow{2}*{0.71403}\\
		~ & calibration & 1.08154 & 0.95360 & 0.97168 & 0.99172 &\textbf{0.98647} & 0.92132& \\
		\hline
		\multirow{2}*{DNN} & AUC & 0.66112 & 0.73587 & 0.73289 &0.67318 & 0.77273 & 0.66772 &\multirow{2}*{0.71225}\\
		~ & calibration & 0.53774 & 0.98387 & 0.89748 & 0.97924 &1.34186 & 0.94484& \\
		\hline
		\multirow{2}*{Wide\&Deep} & AUC & 0.65911 & 0.73539 & 0.73524 &\textbf{0.67611} & 0.77529 & 0.66775&\multirow{2}*{0.71266}\\
		~ & calibration & 0.56710 & 1.02160 & 0.94578 & \textbf{1.00247} &1.39293 &0.90866 & \\
		\hline
		\multirow{2}*{DCN} & AUC & 0.66202 & 0.73557 & 0.73710 &0.67169 & 0.77528 & 0.66723&\multirow{2}*{0.71279}\\
		~ & calibration & 0.55615 & 0.98300 & 0.90693 & 1.05940 &1.36822 & 0.94575& \\
		\hline
		\multirow{2}*{DeepFM} & AUC & 0.65700 & 0.73562 & 0.73587 &0.67219 & 0.77404 & 0.66916&\multirow{2}*{0.71294}\\
		~ & calibration & 0.58993 & 1.01282 & 0.95011 & 1.05969 &1.37286 & 0.96842& \\
		\hline
		\multirow{2}*{DADNN-MLP} & AUC & 0.70451 & 0.73657 & 0.74836 &0.67146 & 0.78617 & 0.66766&\multirow{2}*{0.71808}\\
		~ & calibration & \textbf{1.02730} &  1.01409 & \textbf{1.00267} & 1.02859 &0.89886 & 0.89494& \\
		\hline
		\multirow{2}*{DADNN-MMoE} & AUC & \textbf{0.70462} & \textbf{0.74020} & \textbf{0.75532} &0.67292 & \textbf{0.78711} & \textbf{0.67472}&\multirow{2}*{\textbf{0.72264}}\\
		~ & calibration & 1.04104 &\textbf{0.98831} & 0.97281 & 0.98272 &0.97660 & \textbf{0.97541}& \\
		\hline
	\end{tabular}
	\label{tab:offline_performance_on_industrial_dataset}
\end{table*}

\indent We compare the following approaches with our models:
\begin{itemize}
	\item DNN: The concatenation of all the embedding vectors is fed into multiple layer perceptron. Then, an output layer predicts the probability whether the user will click the ad with a sigmoid function. It acts as a strong baseline for our model comparison.
	\item Wide\&Deep: Wide\&Deep\cite{2016Wide} is a CTR prediction model which considers both memorization and generalization. It contains two parts: wide part for memorization and deep part for generalization.
	\item DCN: Deep\&Cross network\cite{2017Deep} conncatenates a DNN model and a novel cross network to efficiently  capture both the implicit and explicit high-order features interactions.
	\item DeepFM: DeepFM\cite{2017DeepFM} combines the function of factorization machines and deep neural network for feature learning in a network structure.
	\item DADNN-MLP: The DADNN-MLP proposed in this paper. Its shared-bottom block is multilayer perceptron.
	\item DADNN-MMoE: The DADNN-MMoE model uses multi-gate mixture-of-experts to substitute the hard shared-bottom block.
\end{itemize}

\indent Table \ref{tab:offline_performance_on_industrial_dataset} lists the AUC, calibration and GAUC values of the above methods. We also tune and train single-scene models by training a separate model for each scene and report their results. It is observed that Wide\&Deep achieves higher GAUC than DNN. The result indicates that combining a wide component and a deep component for prediction is effective. Similarly, DCN and DeepFM with specially designed structures perform better than Wide\&Deep. DNN(single) improves GAUC obviously due to avoiding domain shift. However, training single-scene models does not take advantage of the rich user and ad information from all scenes. We think this is the reason why some scenes don't perform well. It is worth noting that calibrations of all comparison models except DNN(single) indicates inaccurate CTR prediction, especially in scene 1 and 5. As all we know, the ranking strategy can be adjusted as pCTR$\times$bid across all candidates, thus calibration is a very important metric. The inaccurate estimation of CTR may lead to poor online performance. \\
\begin{table*}[t]
	\centering
	\caption{The performance of different components in DADNN.}
	\begin{tabular}{|c|c|cccccc|c|}
		\hline
		Method &\diagbox{Metric}{Scene}& 1 & 2 & 3 & 4 & 5 & 6 & GAUC \\
		\hline
		\multirow{1}*{BASE(DADNN-MLP)} & AUC & 0.70451 & 0.73657 & 0.74836 &0.67146 & 0.78617 & 0.66766&\multirow{1}*{0.71808}\\
		\hline
		\multirow{1}*{BASE With FT NO-KT} & AUC & 0.70978 & 0.73883 & 0.75554 &0.67775 & 0.78592 & 0.67083&\multirow{1}*{0.72099}\\
		\hline
		\multirow{1}*{MMoE With FT NO-KT} & AUC & 0.71126 & 0.74187 & 0.75490 &0.67973 & 0.78933 & 0.67797&\multirow{1}*{0.72508}\\
		\hline
		\multirow{1}*{BASE With FT} & AUC & 0.70471 & 0.74021 & 0.75346 &0.67365 & 0.78887 & 0.67275&\multirow{1}*{0.72204}\\
		\hline
		\multirow{1}*{MMoE With FT} & AUC & 0.71374 & 0.74558 & 0.76243 &0.67926 & 0.79255 & 0.68071&\multirow{1}*{0.72861}\\
		\hline
	\end{tabular}
	\label{tab:ablation_study}
\end{table*}
\indent Among all the compared methods, our proposed DADNNs almost always achieve the best performance and better calibration in every scene. The other models are well designed to predict CTR in one scene, thus the degradation in performance may be due to domain shift among multiple scenes. Our proposed DADNN-MLP outperforms by 0.00514(relatively 0.721\%) in terms of GAUC over DeepFM. 
A slight improvement in the GAUC is likely to lead to a significant increase in the online performance. We can also see that the AUC improvement is significant on each scene. The results indicate that the proposed model could learn a rich and common representation across scenes, whilst still allowing for discriminative characteristics to be tailored for each individual scene. It is also observed that DADNN-MMoE outperforms DADNN-MLP and achieves 0.00456 absolute GAUC gain and relatively 0.635\%. This is because MMoE explicitly models the differences among scenes while still captures the commonness. There are some highlights in our model which have important effects on the final performance. In order to verify the effectiveness of each module, we design the following experiments.

\subsection{Ablation Study}
\indent Although the strong empirical results are demonstrated, the results presented so far have not isolated the specific contributions from each component of the DADNN. In this section, we verify the effect of each module in our proposed model. Moreover, we introduce the pre-trained embedding layer which is trained by another large scene to initialize model parameters for performance improvement. In order to better understand their relative importance, we perform ablation experiments over DADNN in following ways:1) BASE: DADNN-MLP; 2) With FT: introduce the pre-trained embedding layer to finetune the model; 3)NO-KT: remove the knowledge transfer from DADNN; 4) MMoE: DADNN-MMoE. As shown in Table \ref{tab:ablation_study}, BASE With FT performs better than BASE. It indicates that introducing richer data can further improve model performance. 
We conduct comparative experiments with BASE With FT and BASE-NO-KT With FT, where BASE With FT performs better. It reflects knowledge transfer enhances the opportunity of knowledge sharing among multiple scenes, which improves the performance. 
Compared with BASE With FT, MMoE With FT performs much better, which shows that explicitly modeling the commonness and differences among multiple scenes brings a significant improvement.
The performance of MMoE With FT also indicates that combining the knowledge transfer with the MMoE module for prediction further improves the performance.

\subsection{Hyper-Parameters Investigation}
\begin{figure}[t]
	\centering
	\includegraphics[width=0.4\textwidth]{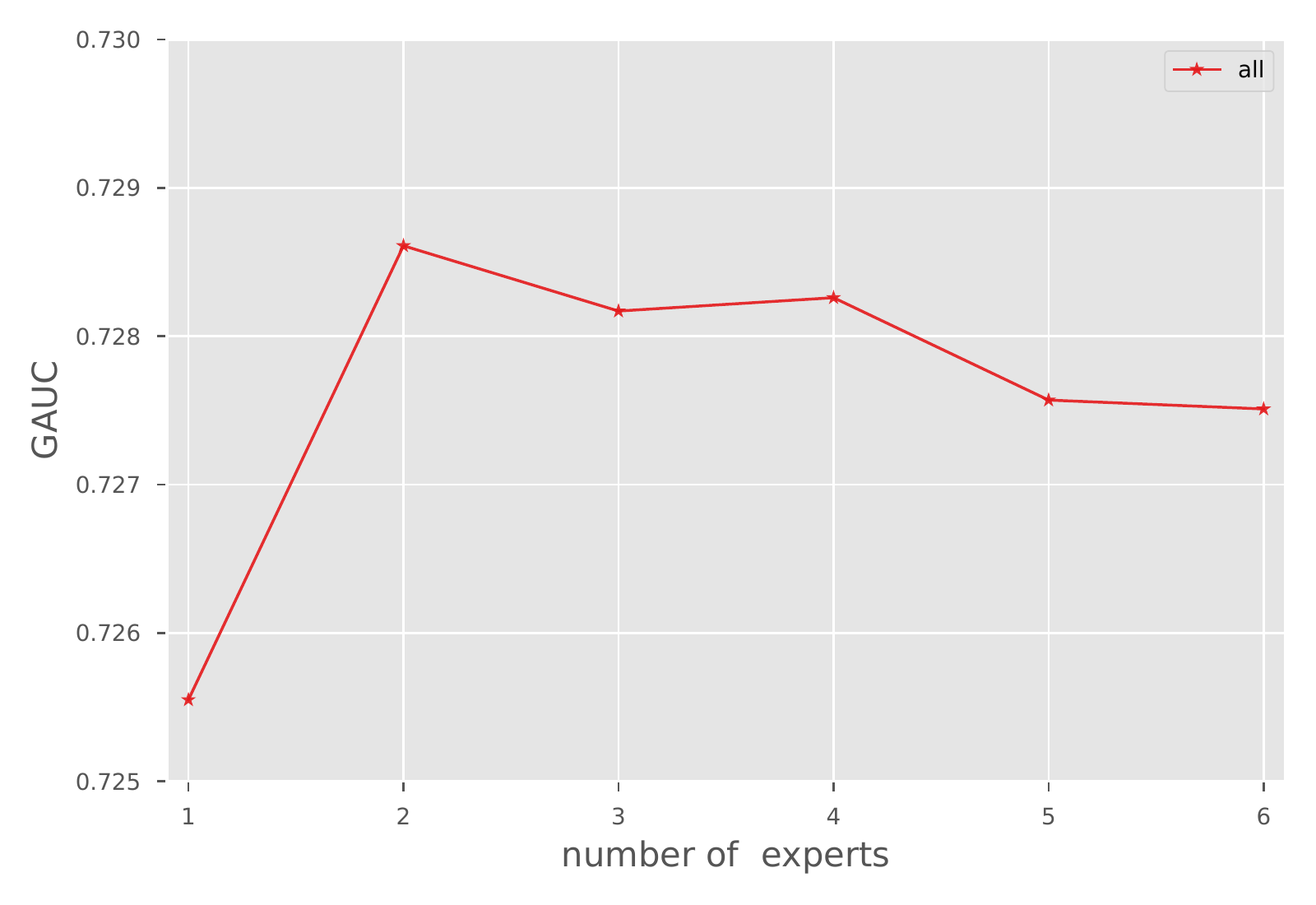}\\
	\caption{The GAUC with different numbers of experts on industrial dataset.}
	\label{fig:gauc_expert}
\end{figure}
\begin{figure}[t]
	\centering
	\includegraphics[width=0.4\textwidth]{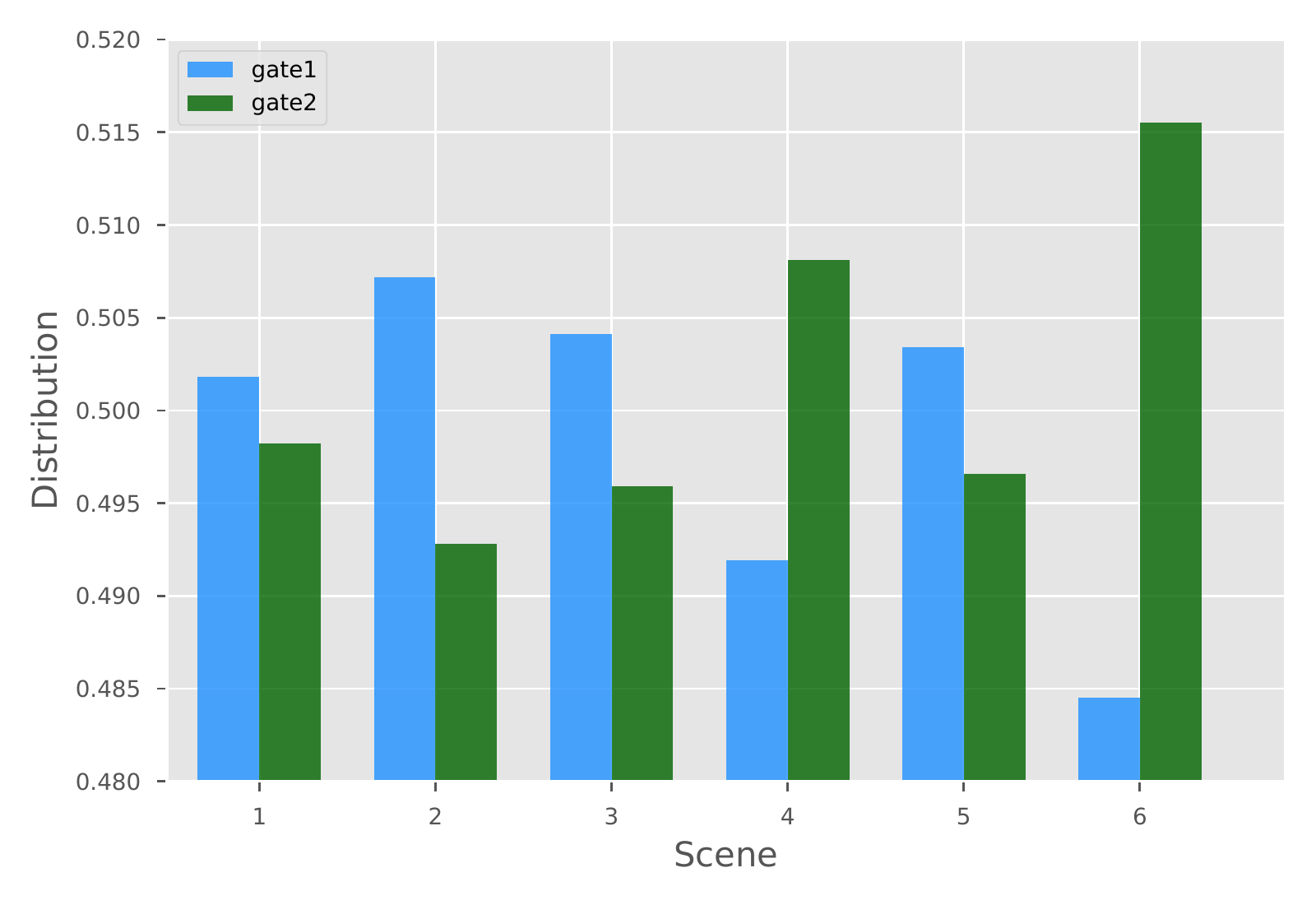}\\
	\caption{Softmax Gate Distribution for multiple scenes.}
	\label{fig:MMoE_gate_score}
\end{figure}
\indent In this subsection, we conduct some hyper-parameter investigations to better understand how the MMoE module work in our model. The test GAUC of DADNN-MMoE vs. the number of experts is shown in Fig. \ref{fig:gauc_expert}. To make the comparison fair, we constrain the number of model parameters by changing number of hidden units per expert. When the number of experts is set to one, the DADNN-MMoE is downgraded to the DADNN-MLP. It is observed that introducing MMoE module can improve the GAUC. However, the benefit diminishes if the number of experts keeps growing, possibly for the reason that less hidden units per expert reduces the representation ability of the shared bottom block. We show gate distribution of each scene in Fig. \ref{fig:MMoE_gate_score}. The number of experts is set to 2. We can see that MMoE can automatically distinguish the differences among these scenes and balance the shared and non-shared parameters.

\subsection{Online A/B Test}
\begin{figure}
	\centering
	\includegraphics[width=0.4\textwidth]{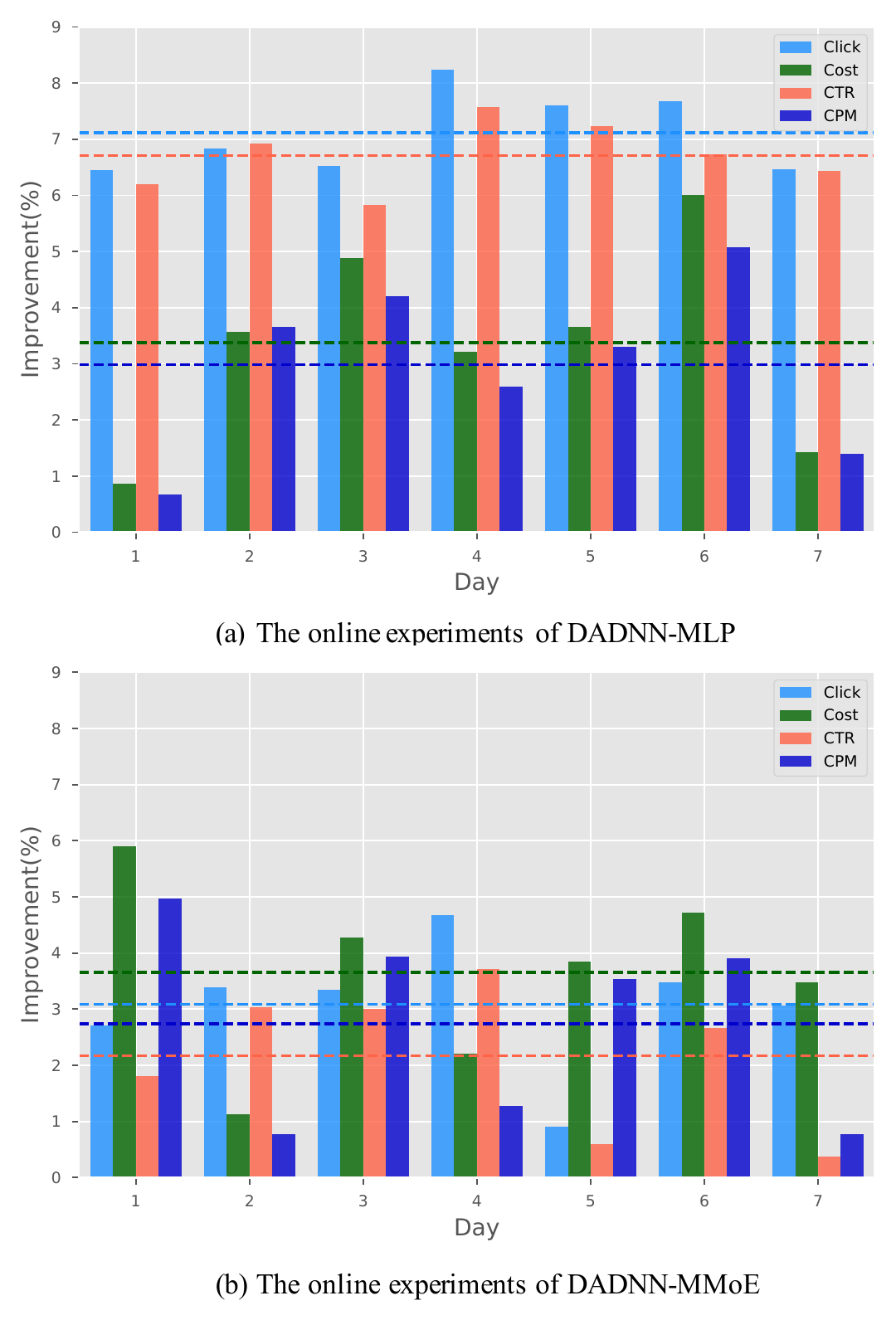}\\
	\caption{Results from online A/B testing}
	\label{fig:Online_performance}
\end{figure}
\indent Extensive online A/B tests have been conducted over two weeks in Jun. 2020 and Aug. 2020 respectively. The first online experiment is shown in Fig. \ref{fig:Online_performance}(a), compared with the baseline model(ie, a well-engineered DCN), the proposed DADNN-MLP, contributes up to 6.7\% CTR , 3.0\% CPM, 7.1\% Clicks, and 3.4\% Cost  promotion. This is a significant improvement and demonstrates the effectiveness of our proposed approach. The second experiment is shown in Fig. \ref{fig:Online_performance}(b), the benchmark model is DADNN-MLP, which is our last online serving model. DADNN-MMoE has improved CTR by 2.2\%, CPM by 2.7\%, Clicks by 3.0\%, and Cost by 3.7\%. The results demonstrate the effectiveness of introducing the MMoE module in practical CTR prediction tasks. DADNN-MMoE has been deployed online and serves the traffic for multiple scenes.

\section{Conclusion}
\indent In this paper, we propose a novel CTR prediction model, namely Domain-Aware Deep Neural Network(DADNN), to serve for multiple scenes of online advertising. Both offline and online experimental results demonstrate the efficiency and effectiveness of DADNN over several state-of-the-art methods. More importantly, our method is highly resource efficient. Specifically, shared-bottom block enables much more expressive combinations of features to be learned for generalisation across scenes. Furthermore, MMoE module explicitly learns both scene-shared and scene-specific representations, which boosts the performance. Then, the proposed routing and domain layer helps to allow for discriminative characteristics to be tailored for each individual scene. At last, knowledge transfer is introduced to promote interactions among multiple scenes through a loss fuction and further improve the performance. In the future, we will investigate more sophisticated architectures and serve an increasing number of scenes. We hope this work could inspire further research efforts into multi-scene modeling.

\bibliographystyle{IEEEtran}
\bibliography{ref}

\end{document}